\documentclass[letterpaper, 10 pt, conference]{ieeeconf}  

\IEEEoverridecommandlockouts                              

\usepackage{amsmath,amsfonts}
\usepackage{algorithmic}
\usepackage{algorithm}
\usepackage{array}
\usepackage[caption=false,font=normalsize,labelfont=sf,textfont=sf]{subfig}
\usepackage{textcomp}
\usepackage{stfloats}
\usepackage{url}
\usepackage{verbatim}
\usepackage{graphicx}
\usepackage{cite}
\usepackage{booktabs}
\usepackage{lipsum} 
\usepackage{multirow}
\usepackage{caption}
\usepackage{newtxmath}

\usepackage[caption=false,font=normalsize,labelfont=sf,textfont=sf]{subfig}

\usepackage{amsmath}

\DeclareMathOperator*{\argmin}{arg\,min}
\newcommand{\wbigcup}{\mathop{\bigcup}\displaylimits}
\usepackage{tikz}
\usetikzlibrary{positioning}

\title{\LARGE \bf
A Spatiotemporal Correspondence Approach to Unsupervised LiDAR Segmentation with Traffic Applications
}

\author{Xiao Li, Pan He$^\dagger$, Aotian Wu, Sanjay Ranka, Anand Rangarajan
\thanks{Xiao Li, Aotian Wu, Sanjay Ranka, and Anand Rangarajan are with the Dept. of Comput. \& Info. Sci. \&  Eng., University of Florida, Gainesville, FL 32611. Pan He is with the Dept. of Comput. Sci. and Soft. Eng., Auburn University, Auburn, AL 36849. $^\dagger$ indicates the corresponding author.}
}

\begin{document}




\maketitle
\thispagestyle{empty}
\pagestyle{empty}

\begin{abstract}
We address the problem of unsupervised semantic segmentation of outdoor LiDAR point clouds in diverse traffic scenarios. The key idea is to leverage the spatiotemporal nature of a dynamic point cloud sequence and introduce drastically stronger augmentation by establishing spatiotemporal correspondences across multiple frames. We dovetail clustering and pseudo-label learning in this work. Essentially, we alternate between clustering points into semantic groups and optimizing models using point-wise pseudo-spatiotemporal labels with a simple learning objective. Therefore, our method can learn discriminative features in an unsupervised learning fashion.  We show promising segmentation performance on Semantic-KITTI, SemanticPOSS, and FLORIDA benchmark datasets covering scenarios in autonomous vehicle and intersection infrastructure, which is competitive when compared against many existing fully supervised learning methods. This general framework can lead to a unified representation learning approach for LiDAR point clouds incorporating domain knowledge.
\end{abstract}


\section{INTRODUCTION}
Semantic LiDAR segmentation is a fundamental problem in computer vision which supports downstream application areas such as autonomous vehicles, robotics, augmented reality, and human-computer interaction. 
To achieve good performance, one popular approach is to use deep neural networks for extracting discriminative features via supervised learning provided we have sufficiently massive data and labels.  However, compared to labeling semantic labels for pixels from images, it is much more time-consuming, challenging, and expensive to annotate pointwise semantic labels for each point cloud that is non-intuitive and sparse, which explains why large-scale 3D datasets for semantic segmentation are scarce. For example, it takes roughly 4.5 hours to label a point cloud covering a small residential area of $100$m by $100$m \cite{semantickitti}. 

Early research works \cite{mei2019semantic,wu2023florida} aim to reduce the annotation cost and effort via semi-supervised learning. They leverage easy-to-acquire unlabeled data for scalable LiDAR segmentation but require considerable pointwise labeled LiDAR data for sufficient learning under supervision. 
Developing efficient and effective unsupervised approaches remains  a challenging problem and may require combining object and group discovery with exhaustive and dense labeling over data elements such as pixels or points following classic Gestalt Principles \cite{koffka2013principles}. 

In this work, we aim to eliminate the need for any pointwise ground truth annotations for LiDAR data. Specifically, we seek to learn pointwise representations or feature embeddings by integrating self-supervised and deep-learning models. We have recently witnessed significant progress in self-supervised and unsupervised learning \cite{uss_mask,salasnext,jsis3d}, where most learning systems have focused on generating a single feature vector for a given sample, e.g., image-level classification or 3D shape classification. However, such systems could rely on only a few salient, distinct, and stable features partially taken from the original data and potentially ignore the remaining features. Ideally, segmenting the LiDAR point cloud must examine every point in terms of whether it comes from the background, stationary objects, dynamic objects, or are noisy, outlier points. Furthermore, significant intraclass variations (e.g., pedestrians at different distances) and interclass similarities (e.g., scooter riders and pedestrians) further prevent accurate segmentation.

In this paper, we develop  a practical unsupervised LiDAR segmentation method by switching between clustering the points into semantic groups based on their feature embeddings and optimizing models using pointwise pseudo-semantic labels generated from clustering. Inspired by \cite{picie}, we learn point embeddings by enforcing equivariance for geometric transformations. It is worth noting that such a framework relies on data augmentation carefully designed to capture the underlying equivariance and preserve the instance or semantic identities between augmented samples. The significant impact of data augmentation strategies has been explored in various methods such as InfoMin \cite{tian2020makes}
for learning better feature representations. In our work, we observe that the spatiotemporal nature of a dynamic point cloud sequence introduces drastically stronger augmentation because their points are constantly changing over time due to object motion, occlusion, and observer movement. Therefore, we propose leveraging spatiotemporal patterns to create stronger augmented views across different frames. To do so, we introduce a straightforward objective that considers the spatiotemporal correspondence as the additional constraint. 

Despite the simplicity of our developed approach, it is competitive with some well-known prior work with fully supervised learning having ground truth labels. A systematic pipeline of LiDAR alignment, multiple dynamic object segmentation and tracking, and correspondence labeling is developed to establish spatiotemporal correspondence in multiple datasets with different types of dataset characteristics. To summarize, we make the following contributions:

\begin{itemize}
    \item We demonstrate that spatiotemporal correspondence can largely help improve the performance of unsupervised LiDAR semantic segmentation.
    \item We develop an unsupervised segmentation framework for LiDAR data from dynamic scenes.
    \item To the best of our knowledge, we are the first to conduct unsupervised LiDAR semantic segmentation in autonomous driving and infrastructure datasets.
    \item We show promising segmentation performance on Semantic-KITTI, SemanticPOSS, and FLORIDA benchmark datasets.
\end{itemize}

The rest of the paper is structured as follows. Section~\ref{relwork} presents the related work. The methodology developed for unsupervised segmentation is described in Section~\ref{methodology}. Section~\ref{expt} outlines the conducted experiments. Conclusions are described in Section~\ref{conclusion}.

\section{RELATED WORK}\label{relwork}
\noindent \textbf{Point Cloud Models}. Point clouds are unordered and challenging to process with standard convolutional neural networks. 
PointNet \cite{pointnet} was the first network that operated on raw point clouds by addressing point cloud data's unordered format and invariance. PointNet++ \cite{pointnet2} follows the idea of PointNet and proposes to utilize both global information and local details with the farthest sampling layer and a grouping layer. Instead of consuming point cloud data directly, the pioneering SqueezeSeg \cite{wu2018squeezeseg} projects the data onto a 2D image using spherical range projection and processes the projected 2D image with an encoder-decoder architecture. Extensions of these works include SqueezeSegV2 \cite{wu2019squeezesegv2}, Rangenet++\cite{milioto2019rangenet++},  SalsaNet \cite{aksoy2020salsanet}, and SalsaNext \cite{salasnext}. We adopted SalsaNext as the backbone as it provides a  good trade-off between speed and accuracy.

\noindent \textbf{Self-supervised Feature Learning}. Self-supervised feature learning aims to extract meaningful visual features without ground truth labels. Substantial research has focused on optimizing specific surrogate tasks such as denoising \cite{vincent2008extracting}, inpainting \cite{pathak2016context}, rotation \cite{gidaris2018unsupervised}, or contrastive learning over multiple augmentations \cite{oord2018representation}. Contrastive learning approaches seek to learn feature representations by contrasting similar and dissimilar pairs of samples via data augmentations. They minimize the distances between positive pairs (i.e., one image and its augmented image) and maximize the distances between negative pairs (i.e., one image and another randomly sampled image).  Recently, noncontrastive self-supervised learning (NC-SSL) methods learn meaningful feature representations using only positive pairs, different from the contrastive approaches using both positive and negative pairs. Despite the lack of negative pairs, NC-SSL approaches use techniques such as extra predictors, stop gradient, batch normalization, decorrelation, whitening, and centering to avoid model collapse and have achieved comparable or better performance. In addition, there are attempts to provide theoretical studies for NC-SSL  \cite{tian2021understanding,wang2021towards}.

\noindent \textbf{Unsupervised Semantic Segmentation}. Deep learning-based semantic segmentation has been widely studied \cite{uss_2d_1}. Most studies focus on supervised learning, while unsupervised semantic segmentation has not been fully explored. The pioneering IIC \cite{iic} extends mutual information-based clustering to pixel-level clustering by outputting a probability map over image pixels. Contrastive Clustering \cite{li2021contrastive} and SCAN \cite{van2020scan} have further improved IIC's results by incorporating negative samples and nearest neighbors as supervision. However, it should be noted that these methods focus on image clustering and do not address the task of semantic segmentation. PiCIE \cite{picie} proposes incorporating geometric consistency as an inductive bias to learn invariance and equivariance for photometric and geometric variations. 
Our work builds a baseline approach by adapting PiCIE for unsupervised LiDAR point cloud segmentation and introduces stronger supervision signals by finding spatiotemporal correspondences from sequences.

\noindent \textbf{Dynamic Object Segmentation}. The dynamic object segmentation approaches can be divided into two categories, namely map-based and map-free approaches. In map-based approaches, they segment dynamic objects from LiDAR point clouds by exploiting prior information from the scene, such as a prebuilt 3D environmental map, to detect and track moving objects by comparing the current LiDAR data against the map \cite{pfreundschuh2021dynamic}. In contrast, map-free approaches rely solely on raw point clouds to perform segmentation. One typical map-free approach is clustering-based segmentation to group points in the point cloud belonging to the same object based on their point or feature similarity. One popular clustering-based algorithm is DBSCAN (Density-Based Spatial Clustering of Applications with Noise) \cite{campello2013density}, which partitions data points that are densely packed into the same group while marking data points that lie in sparser regions as noise. Other clustering-based approaches includ DBSCAN \cite{dbscan} and K-means clustering.



\section{METHODOLOGY}\label{methodology}

\begin{figure*}
    \centering    \subfloat{\includegraphics[width=\textwidth]{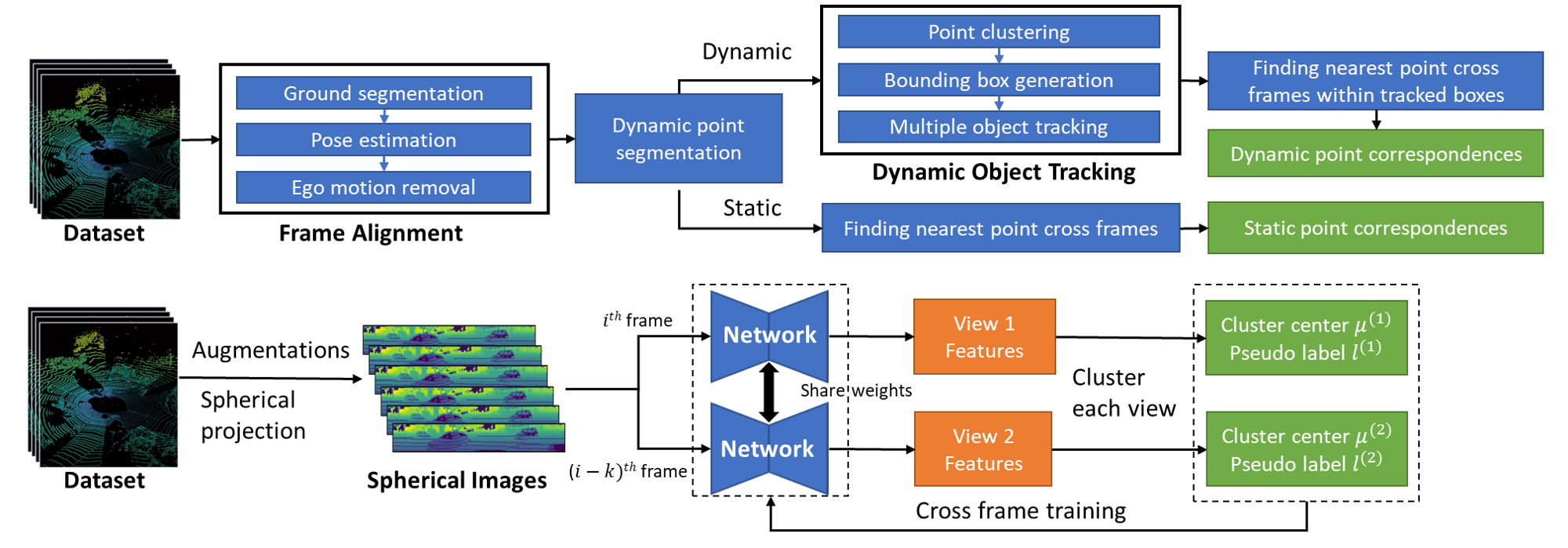}}
    \caption{Method overview: Finding correspondences for static objects and dynamic objects (top). Feature clustering and model training (bottom)}
    \label{fig:main_fig}
\end{figure*}

We aim to segment a point cloud into semantic objects or regions without providing ground-truth semantic or instance labels. Formally, given a 3D point cloud (e.g., obtained from a LiDAR sensor) denoted as $\boldsymbol{P} = \{\boldsymbol{p}_1, \boldsymbol{p}_2, ..., \boldsymbol{p}_n\}$, where each point $\boldsymbol{p}_i$ is represented by its 3D coordinates $(x_i, y_i, z_i)$ and its associated feature (e.g., intensity), the goal of unsupervised segmentation is to find a partition of $\boldsymbol{P}$ into $m$ segments $S = \{S_1, S_2, ..., S_m\}$, such that: (i) Each segment $S_j (1 \leq j \leq m)$ represents a coherent region of points that belongs to the same semantic class or the same instance; (ii) The segments are mutually exclusive and exhaustive such that every point in $P$ belongs to exactly one segment, and the segments cover all points in $\boldsymbol{P}$.

\subsection{Baseline Approach} \label{sec:baseline}

\subsubsection*{Input Representation}

Inspired by \cite{wu2018squeezeseg,salasnext}, we prefer the compact and efficient Range View (RV)-based representation for LiDARs that projects a 3D LiDAR point cloud onto a spherical surface. Formally, each point $\boldsymbol{p}_i$ is projected to an image coordinate $(u_i,v_i)$ as follows: 
\begin{equation}
    \begin{pmatrix}
        u_i\\
        v_i\\
    \end{pmatrix}  =
     \begin{pmatrix}
        \frac{1}{2} \left [ 1 - \arctan(y_i, x_i) \pi^{-1} \right ] W \\
        \left [ 1- (\arcsin(\boldsymbol{z}_i/r_i)) + f_{\mathrm{down}})f^{-1} \right]H\\
    \end{pmatrix}
\end{equation}
where $r_i = \sqrt{x_i^2+y_i^2+z_i^2}$ is the range value of each point $\boldsymbol{p}_k$. $H$ and $W$ denote the height and the width of the projected 2D RV image, respectively. $f = |f_{\mathrm{down}}| +  |f_{\mathrm{up}}|$ defines the sensor vertical field of view. During projection, the 3D point coordinate $(x_i, y_i, z_i)$, its intensity value ${intensity}_i$, and the range value $r_i$ are stored in separate channels, generating a $\left [H \times W \times 5 \right ]$ RV image.  

\subsubsection*{Backbone Network} 

The backbone network is the popular SalsaNext \cite{salasnext}, an upgraded version of SalsaNet \cite{aksoy2020salsanet}, featuring an encoder-decoder architecture. With its model parameters $\boldsymbol{w}$, we project the RV image representation of size $H \times W \times 5$ to a feature map $\boldsymbol{F}_{\boldsymbol{w}}$ of size $H \times W \times C$.

\subsubsection*{Unsupervised Segmentation}


Assume we have a set of unlabeled point clouds $\boldsymbol{P}^{j}, j = 1, . . . , n$, and their corresponding feature representations $\boldsymbol{F}_{\boldsymbol{w}}^{j},  j = 1, . . . , n$, after deploying the SalsaNext network. We denote $\boldsymbol{f}_{\boldsymbol{w}}^{j}(u_i,v_i) \in \boldsymbol{F}_{\boldsymbol{w}}^{j}$ as the feature map of the $j$-th point cloud at the image coordinate $(u_i, v_i)$. The baseline approach follows  PiCIE \cite{picie} and alternates between two key steps:

\begin{enumerate}
    \item Apply K-means to cluster current feature embeddings:
    \begin{equation}
        \min_{
            \boldsymbol{\mu}, \boldsymbol{l}}
            \sum_{i,j}   \left\|
                            \boldsymbol{f}_{\boldsymbol{w}}^{j}(u_i,v_i) - \boldsymbol{\mu}_{\boldsymbol{l}_{j,i}}
                       \right\|^2_{2}
    \end{equation}
    where $\boldsymbol{l}_{j,i}$ denotes the cluster label of the pixel at the location $(u_i, v_i)$ in the RV feature map of the $j$-th point cloud, and $\boldsymbol{\mu}_k$ is the $k$-th cluster centroid.
    \item Use the assigned cluster labels to update the model parameters based on the following loss:
    \begin{flalign}
        &\min_{
            {\boldsymbol{w}}}
            \sum_{i,j}   \ell_{CE}(\boldsymbol{f}_{\boldsymbol{w}}^{j}(u_i,v_i), \boldsymbol{l}_{j,i}, \boldsymbol{\mu}),\quad \mathrm{where} \\
        & \ell_{CE}(\boldsymbol{f}_{\boldsymbol{w}}^{j}(u_i,v_i), \boldsymbol{l}_{j,i}, \boldsymbol{\mu}) = - \log \frac{e^{-d(\boldsymbol{f}_{\boldsymbol{w}}^{j}(u_i,v_i),\boldsymbol{\mu}_{\boldsymbol{l}_{j,i}}) }}{\sum_k e^{-d(\boldsymbol{f}_{\boldsymbol{w}}^{j}(u_i,v_i), \boldsymbol{\mu}_k)}}
        \label{eq:lCE}
    \end{flalign}
    and $d(\cdot,\cdot)$ denotes the cosine distance. Equation~(\ref{eq:lCE}) defines a nonparametric prototype-based classifier to label pixels based on their distances from the centroids.
\end{enumerate}

We adapt these two key steps for learning  equivalence with respect to geometric transformations on point clouds. We apply several augmentations to point clouds before projecting them to RV images. Then, we create two views of the same point cloud $\boldsymbol{P}^{j}$ via two geometric transformations, $G_1$ and $G_2$, and generate two feature maps for $\boldsymbol{p}_i$ in $\boldsymbol{P}^{j}$:
\begin{flalign}
       & \boldsymbol{z}^{(1)}_{j, i} =  \boldsymbol{f}_{\boldsymbol{w}}^{j}(u_i,v_i, G_1), \\
       & \boldsymbol{z}^{(2)}_{j, i} = \boldsymbol{f}_{\boldsymbol{w}}^{j}(u_i,v_i, G_2).
\end{flalign}
We then separately perform clustering on the two views to obtain two separate sets of pseudo-labels and centroids:
\begin{flalign}
       & \boldsymbol{l}^{(1)}, \boldsymbol{\mu}^{(1)} = \argmin_{
            \boldsymbol{\mu}, \boldsymbol{l}}
            \sum_{i,j}   \left\|
                            \boldsymbol{z}^{(1)}_{j, i} - \boldsymbol{\mu}_{\boldsymbol{l}_{j,i}}
                       \right\|^2_2 \label{eq:cluster_1} \\
       &\boldsymbol{l}^{(2)}, \boldsymbol{\mu}^{(2)} = \argmin_{
            \boldsymbol{\mu}, \boldsymbol{l}}
            \sum_{i,j}   \left\|
                            \boldsymbol{z}^{(2)}_{j, i} - \boldsymbol{\mu}_{\boldsymbol{l}_{j,i}}
                       \right\|^2_2 \label{eq:cluster_2}
\end{flalign}
Following PiCIE \cite{picie}, we use two loss functions to ensure the consistency of embeddings within each view and across two views:
\begin{flalign}
E_{\mathrm{within}}({\boldsymbol{w}}) =  & \sum_{i,j}   \ell_{CE}(\boldsymbol{f}_{\boldsymbol{w}}^{j}(u_i,v_i, G_1), \boldsymbol{l}^{(1)}_{j,i}, \boldsymbol{\mu}^{(1)}) \nonumber + \\
&
     \sum_{i,j}   \ell_{CE}(\boldsymbol{f}_{\boldsymbol{w}}^{j}(u_i,v_i, G_2), \boldsymbol{l}^{(2)}_{j,i}, \boldsymbol{\mu}^{(2)}), \\
   E_{\mathrm{cross}}({\boldsymbol{w}}) =  & \sum_{i,j}   \ell_{CE}(\boldsymbol{f}_{\boldsymbol{w}}^{j}(u_i,v_i, G_1), \boldsymbol{l}^{(2)}_{j,i}, \boldsymbol{\mu}^{(2)})  \nonumber + \\
&
     \sum_{i,j}   \ell_{CE}(\boldsymbol{f}_{\boldsymbol{w}}^{j}(u_i,v_i, G_2), \boldsymbol{l}^{(1)}_{j,i}, \boldsymbol{\mu}^{(1)}), \\
   E_{\mathrm{total}}({\boldsymbol{w}}) =  & E_{\mathrm{within}}({\boldsymbol{w}}) + E_{\mathrm{cross}}({\boldsymbol{w}}),
\label{eq:picieloss}
\end{flalign}
which encourages feature equivariance to geometric transformations.

\subsection{Spatiotemporal Learning via Auto Labeling}

\subsubsection*{LiDAR Alignment}
Although we may have ground-truth pose for LiDAR sequences, we prefer a flexible approach wherein we use raw sequential LiDAR sequences as the input and estimate relative pose between two LiDAR scans on-the-fly via popular registration algorithms such as  Iterative Closest Point (ICP) \cite{chen1992object}. Specifically, given a LiDAR point cloud sequence $\boldsymbol{P}^{1:T}$ of length $T$, we apply the following data processing pipeline to obtain the estimated poses.  First, for each point cloud $\boldsymbol{P}^{(t)} \in \boldsymbol{P}^{1:T}$, we remove its ground points using a fast ground segmentation algorithm \cite{lim2021patchwork}. Second, we use the Statistical
Outlier Removal (SOR) Filter \cite{rusu2007towards} to remove the outliers. Finally, we align all point clouds w.r.t.  the first point cloud $\boldsymbol{P}^{t=1}$ using ICP  \cite{chen1992object}, which provides the estimated ego-motion matrix. We now obtain a LiDAR point cloud sequence aligned to $\boldsymbol{P}^{t=1}$ denoted as $\boldsymbol{P}^{1:T}_{\mathrm{aligned}}$.

\subsubsection*{Dynamic Object Segmentation}
\label{dynamicSeg}
In a real-world environment, static objects and background points only have ego-motion; therefore, identifying static objects and background regions and finding the correspondences between them is relatively easy via simple nearest neighbor searches between two aligned LiDAR point clouds. For example, suppose the distances of one point to its nearest neighbor points in the other aligned point clouds are small. In that case, it tends to be a stationary point from static objects or the background. However, dynamic objects have extra object motion, which means points on them cannot easily establish correspondences using nearest neighbor searches due to potentially significant motion or occlusion.

We identify potentially dynamic objects based on the observation that a dynamic object will not appear in the same location in all aligned LiDARs due to object motion. Therefore, the  distances for each point in the dynamic object in one point cloud $\boldsymbol{P}^{i}_{\mathrm{aligned}}$ to its nearest neighbor points in other reference point clouds are likely to be higher than those points in static objects or the background.  Formally, given the timestep set $T_{\mathrm{ref}} = \{i-M:i-1\} \cup \{i+1: i+M\}$, we denote all reference point clouds w.r.t. $\boldsymbol{P}^{i}_{\mathrm{aligned}}$ as $\boldsymbol{P}^{\mathrm{ref}} = \{ \boldsymbol{P}^t_{\mathrm{aligned}} \}_{t \in T_{\mathrm{ref}}}$ and define the following rule to compute the dynamic score of each point and identify the dynamic object points. 
\begin{equation}
    \label{eq:dynamic2}
    \mathrm{score}_i = 1 - e^{- \lambda \max\limits_{t \in T_{\mathrm{ref}}} \| \boldsymbol{p}_i - \boldsymbol{p}_{i,t}\|_{2} }
\end{equation}
\begin{equation}\label{eq:dynamic}
     \boldsymbol{p}_i \in
\begin{cases}
     \text{dynamic} ,& \text{if } \mathrm{score}_i   \geq \epsilon   \\ 
    \text{non-dynamic},              & \text{otherwise}
\end{cases}
\end{equation}
where $\boldsymbol{p}_{i,t}$ denotes the closest point to $\boldsymbol{p}_i$  on the aligned point cloud $\boldsymbol{P}^{t}_{\mathrm{aligned}}$. The number of frames we take into consideration to compute the dynamic score is $2M$, and $\lambda$ is a hyperparameter for controlling sensitivity.  The function in Equation~\eqref{eq:dynamic} ensures that its output values are between $[0,1)$. A higher output value means a higher chance of being the points from a dynamic object. 
We introduce a threshold $\epsilon$ such that the point whose output value is greater than $\epsilon$ belongs to a dynamic object.

\subsubsection*{Multiple Dynamic Object Clustering}
\label{sec:dynamic_cluster}
We choose DBSCAN \cite{dbscan} to group these dynamic points into instances.
For each point cloud in the aligned point cloud set $\boldsymbol{P}^{1:T}_{\mathrm{aligned}}$, we first remove all the non-dynamic points. We then augment each remaining point with the $\mathrm{score}_i$ and use DBSCAN to generate a set of dynamic object clustering:
\begin{equation}
    \boldsymbol{S}_{\mathrm{dynamic}} = \wbigcup_{k \in \{1, \cdots, K\}} \boldsymbol{s}_k
\end{equation}
where $\boldsymbol{s}_k$  denotes the $k$-th clustering step.

Based on the clustering results, we generate a bounding box $\boldsymbol{b}_k = (c_k, \gamma, l, w, h, v)$ for each cluster $\boldsymbol{S}_k$ where $c_k \in \mathrm{R}^3$ is the center coordinates; $l, w, h$ denote the length, width, and height of the bounding box, respectively. $v$ is the volume of a bounding box where $v=l×w×h$, $\gamma$ is the heading angle. We remove a bounding box that is either too small or too large to consider outlier segments. For example, a bounding box containing less than $N_{min}$ points or having a side length greater than a size threshold. Finally, we can obtain a set of bounding box instances $\boldsymbol{B} = \{\boldsymbol{b}_m\}_{m=1}^{K_b}$ with $K_b \leq K$. 

\subsubsection*{Multiple Dynamic Object Tracking}

Our next step is to track instance bounding boxes such that we can associate all dynamic object instances across multiple frames to establish the correspondence. To do so, we construct a cost matrix $\boldsymbol{C} \in \mathrm{R}^{K_b^t \times K_b^{t-1}}$ between all box instances $\boldsymbol{B}^t$ at  time $t$ and all previous tracked box instances $\boldsymbol{B}^{t-1}$ at time $t-1$. We then formulate the association problem of instances as a bipartite graph-based linear assignment problem, which can be solved with Jonker-Volgenant algorithm \cite{crouse2016implementing}. Following \cite{chen2022automatic}, we construct the cost matrix based on instance similarity w.r.t. three geometric features, namely the center distance, the overlapping volume between bounding boxes, and the change of volume between each pair of instances. 
Using this simple tracking approach, we obtain the box-level correspondence between dynamic object instances in different frames.

\subsubsection*{Auto Correspondence Labeling}
\label{auto_correp}
Our final auto correspondence strategy for static, dynamic objects, and background is therefore formulated as follows:
\begin{enumerate}
    \item For points from static objects and background, we directly find each point's correspondence in other point clouds by finding the closest point in the aligned point clouds (if their distance is smaller than a threshold).
    \item For points from dynamic objects, we find each point's correspondence by finding its associated bounding box and bounding boxes in other point clouds with the same instance ID. We then apply ICP between all points within the point's bounding box and other boxes to establish the point-to-point correspondences based on a simple rigid motion assumption.
\end{enumerate}

\subsubsection*{Spatiotemporal Learning} \label{sec:spatiotemporal}

After establishing the correspondences,  we follow the PiCIE learning pipeline for unsupervised feature learning. The critical difference is that we conduct the feature learning in a spatiotemporal fashion, which is considered a stronger augmentation strategy than the two-view augmentation. Formally, given two point clouds $\boldsymbol{P}^t, \boldsymbol{P}^{t-k}$ at different timestamps $t$ and $t-k$, we generate the feature maps for each point $\boldsymbol{p}_i$ in the
point cloud $\boldsymbol{P}^{t}$ and each point $\boldsymbol{p}_j$ in the
point cloud $\boldsymbol{P}^{t-k}$:
\begin{flalign}
       & \boldsymbol{z}_{t,i} =  \boldsymbol{f}_{\boldsymbol{w}}^{t}(u_i,v_i, G_1) \\
       & \boldsymbol{z}_{t-k,j} = \boldsymbol{f}_{\boldsymbol{w}}^{t-k}(u_j,v_j, G_{2}).
\end{flalign}

Similar to Equations~\eqref{eq:cluster_1} and \eqref{eq:cluster_2}, we obtain two separate sets of pseudo-labels and centroids for $\boldsymbol{P}^t, \boldsymbol{P}^{t-k}$, which are denoted as $\boldsymbol{l}^{(1)},\boldsymbol{\mu}^{(1)}$ and $ \boldsymbol{l}^{(2)},\boldsymbol{\mu}^{(2)}$, respectively.  For each point $\boldsymbol{p}_i$ in the
point cloud $\boldsymbol{P}^{t}$, we denote its corresponding point in  the
point cloud $\boldsymbol{P}^{t-k}$ as the point $\boldsymbol{p}_{i \rightarrow j}$. We then introduce additional spatiotemporal consistency of embeddings between them to encourage  the feature vector of $\boldsymbol{p}_i$ to match the cluster labels
and centroids of $\boldsymbol{p}_{i \rightarrow j}$.
Using the point-wise correspondences, we map the pseudo-label of one view to another, obtaining another two sets of pseudo-labels $\boldsymbol{l}^{(2 \rightarrow 1)}$ and $\boldsymbol{l}^{(1 \rightarrow 2)}$. Formally, the spatiotemporal loss is defined as
\begin{flalign}
    E_{ST}({\boldsymbol{w}}) = &  \sum_{i,t}   \ell_{CE}(\boldsymbol{f}_{\boldsymbol{w}}^{t}(u_i,v_i, G_1),  \boldsymbol{l}^{(2 \rightarrow 1)}_{t-k, i}, \boldsymbol{\mu}^{(2)})  + \nonumber  \\
    & \sum_{j,t}   \ell_{CE}(\boldsymbol{f}_{\boldsymbol{w}}^{t-k}(u_j,v_j, G_{2}),  \boldsymbol{l}^{(1 \rightarrow 2)}_{t,j}, \boldsymbol{\mu}^{(1)})
\label{eq:stloss}
\end{flalign} 
where $E_{ST}$ denotes a spatiotemporal loss. We randomly select two frames with an interval of $n$ uniformly picked from the set $F  = \{5, 10,15, 20,25, 30\}$.

\subsubsection*{Discriminative Push-Pull Loss Function}

In this work, we customize another discriminative loss function that was proposed in \cite{jsis3d} as a regularization. The key idea of the original loss function is to pull features that belong to the same group (e.g., same class) together and push features that belong to different groups away from each other for learning discriminative feature representations. 
As we cluster a point cloud into a set of clusters using DBSCAN, we customize the original design by defining the same group based on the size of their bounding box, observing that point cloud clusters that belong to the same semantic class tend to have a similar bounding box size. Hence, we define three semantic groups, namely small dynamic group, large static group, and ground group. The integrated loss function is 
\begin{equation}
    E_{\mathrm{final}}({\boldsymbol{w}}) = \alpha E_{\mathrm{within}}({\boldsymbol{w}}) + \beta E_{ST}({\boldsymbol{w}}) + \gamma E_{\mathrm{dloss}}({\boldsymbol{w}})
\end{equation}
where $E_{\mathrm{within}}$ ensures the consistency of embeddings within each view, $E_{ST}$ is the proposed spatiotemporal loss for ensuring additional spatiotemporal consistency cross frames, and $E_{\mathrm{dloss}}$ is the customized discriminative loss for feature regularization.

\section{EXPERIMENTAL RESULTS}\label{expt}

We now provide detailed experimental results on  a variety of public domain and our own datasets and conduct a suitable ablation study.

\begin{table*}[t!]
\centering
\caption{Comparison of different methods on Semantic-KITTI dataset. }
\Large
\resizebox{\textwidth}{!}{  
\begin{tabular}{ l c c c c c c c c c c c c c c c c c c c c } 
\hline
\toprule
Method & 
\rotatebox[origin=c]{90}{{car}} &
\rotatebox[origin=c]{90}{{bicycle}} &
\rotatebox[origin=c]{90}{{motorcycle}} &
\rotatebox[origin=c]{90}{{truck}} &
\rotatebox[origin=c]{90}{{other-vehicle}} &
\rotatebox[origin=c]{90}{{person}} &
\rotatebox[origin=c]{90}{{bicyclist}} &
\rotatebox[origin=c]{90}{{motorcyclist}} &
\rotatebox[origin=c]{90}{{road}} &
\rotatebox[origin=c]{90}{{parking}} &
\rotatebox[origin=c]{90}{{sidewalk}} &
\rotatebox[origin=c]{90}{{other-ground}} &
\rotatebox[origin=c]{90}{{building}} &
\rotatebox[origin=c]{90}{{fence}} &
\rotatebox[origin=c]{90}{{vegetation}} &
\rotatebox[origin=c]{90}{{trunk}} &
\rotatebox[origin=c]{90}{{terrain}} &
\rotatebox[origin=c]{90}{{pole}} &
\rotatebox[origin=c]{90}{{traffic-sign}} &
\text{mIoU}\\

\midrule
Pointnet \cite{pointnet}  &46.3&1.3&0.3&0.1&0.8&0.2&0.2&0.0&61.6&15.8&35.7&1.4&41.4&12.9&31.0&4.6&17.6&2.4&3.7&14.6
\\
Pointnet++ \cite{pointnet2}  &53.7&1.9&0.2&0.9&0.2&0.9&1.0&0.0&72.0&18.7&41.8&5.6&62.3&16.9&46.5&13.8&30.0&6.0&8.9&20.1\\
SqueezeSeg \cite{wu2018squeezeseg}  &68.8&16.0&4.1&3.3&3.6&12.9&13.1&0.9&85.4&26.9&54.3&4.5&57.4&29.0&60.0&24.3&53.7&17.5&24.5&29.5\\
SalsaNet \cite{aksoy2020salsanet}  &87.5&26.2&24.6&24.0&17.5&33.2&31.1&8.4&89.7&51.7&70.7&19.7&82.8&48.0&73.0&40.0&61.7&31.3&41.9&45.4\\ 
SalsaNext \cite{salasnext} &91.9&48.3&38.6&38.9&31.9&60.2&59.0&19.4&91.7&63.7&75.8&29.1&90.2&64.2&81.8&63.6&66.5&54.3&62.1&59.5\\ 

\midrule

Baseline (Ours) &25.3&0.3&0.1&0.1&0.2&0.1&0.0&0.0&53.8&0.9&28.0&0.1&11.8&4.3&10.7&1.2&24.8&0.8&0.3&8.6\\ 
Baseline + Ego (Ours)& 47.0&0.2&0.0&0.3&0.4&0.3&0.1&0.0&65.1&3.7&27.3&1.3&20.1&8.9&15.2&0.6&22.0&0.4&2.5&11.3 \textcolor{red}{(+2.7)}\\ 

Baseline + ST (Ours)&61.7&0.0&0.0&0.9&0.6&0.2&0.0&0.0&71.7&4.1&19.4&1.4&25.7&10.6&14.9&2.2&32.8&0.5&0.0&13.0 \textcolor{red}{(+4.4)}\\
Baseline + ST + DLoss (Ours) & 63.7&0.4&0.2&0.0&0.7&0.1&0.0&0.0&67.4&6.5&23.1&1.0&26.4&10.4&19.1&1.8&32.4&0.4&0.0&13.4 \textcolor{red}{(+4.8)}\\
\bottomrule

\end{tabular}
}

\label{tab:main_results}
\end{table*}

\subsection{Experimental Setup}
We trained and evaluated our model on the following data sets: Semantic-KITTI \cite{semantickitti}, Semantic-POSS \cite{pan2020SemanticPOSS}, and FLORIDA \cite{wu2023florida}, which cover scenarios in autonomous driving and infrastructure, i.e., traffic intersections.  
We implemented our model in PyTorch and trained the model for 100 epochs with a batch size of 12, using an Adam optimizer with a learning rate of $0.05$. At the 40th epoch, we reduced the learning rate by a factor of $10$.
We consider recent state-of-the-art point-based supervised approaches (PointNet \cite{pointnet} and PointNet++ \cite{pointnet2}) as competitive baselines. Besides, we compare against RV-based supervised approaches, including SqueezeSeg \cite{wu2018squeezeseg}, SalsaNet \cite{aksoy2020salsanet}, SalsaNext \cite{salasnext}, SqueezeSegV2 \cite{wu2019squeezesegv2}, RangeNet++ \cite{milioto2019rangenet++}, UnpNet\cite{li2021unpnet}, and MINet \cite{minet}.
We adopt the mean Jaccard Index or intersection-over-union ($\text{mIoU}$) to evaluate the performance of our method.
$\text{mIoU}$ can be expressed as
\begin{equation}
    \text{mIoU} = \frac{1}{C}\sum_{i=1}^{C}\frac{\text{TP}_i}{\text{TP}_i+\text{FP}_i+\text{FN}_i},
\end{equation}
where $\text{TP}_i$, $\text{FP}_i$, $\text{FN}_i$ denote the number of true positive, false positive, and false negative predictions for class $i$ and 
$C$ is the number of classes.

    

\begin{table*}[h!]
\centering

\caption{Comparison of different methods on SemanticPOSS dataset. }
\begin{tabular}{ c c c c c c c c c c c c c c c c } 
\hline
\toprule
Method & 
people &
rider &
car &
traffic sign &
truck &
plants &
pole&
fence &
building &
bike &
road&
\text{mIoU}\\

\midrule

PointNet++\cite{pointnet2}   &20.8&0.1&8.9&21.8&4.0&51.2&3.2&6.0&42.7&0.1&62.2&20.1\\
SequeezeSegV2\cite{wu2019squeezesegv2}&18.4&11.2&34.9&11.0&15.8&56.3&4.5&25.5&47.0&32.4&71.3&29.8\\
RangeNet++ \cite{milioto2019rangenet++}&14.2&8.2&35.4&6.8&9.2&58.1&2.8&28.8&55.5&32.2&66.3&28.9\\
UnpNet\cite{li2021unpnet}&17.7&17.2&39.2&9.5&13.8&67.0&5.8&31.1&66.9&40.5&68.4&34.3\\
MINet\cite{minet}&20.1&15.1&36.0&15.5&23.4&67.4&5.1&28.2&61.6&40.2&72.9&35.1\\

\midrule     

Baseline (Ours)&  5.4 &      0.8 &      5.7 &    0.4 &   1.0  &     35.9 &     0.5  &  2.6 &   18.9  &         3.9  &    44.7 &10.9   \\ 
Baseline + Ego (Ours) &  0.7  &     0.2  &     19.3  &    0.7  & 4.5 & 21.5 &0.7 &  0.0 & 27.6 & 19.8  & 66.1& 14.7 \textcolor{red}{(+3.8)}\\
Baseline + ST (Ours)  & 2.8  &     2.2 &     27.1 & 0.6  & 3.7 &   25.1 & 0.2  &2.9& 39.6 & 19.4  & 63.3 & 17.0  \textcolor{red}{(+6.1)}\\
Baseline + ST + DLoss (Ours) & 14.6  & 0.4 & 23.8 & 0.9 & 0.0 & 34.7 & 0.1 & 7.1 & 21.0 & 36.6 & 80.8 & 20.0 \textcolor{red}{(+9.1)}\\


\bottomrule

\end{tabular}
\label{tab:semposs}
\end{table*}

\subsection{Experimental Results}



As shown in Table~\ref{tab:main_results}, we reported the performance of our developed models on the Semantic-KITTI dataset. The `baseline' is the model trained on two-view augmentation of the single frame described in Section \ref{sec:baseline}. Both `Ego' and `ST' use the spatiotemporal loss described in Section~\ref{sec:spatiotemporal}. The difference is that we only use static point correspondences for the `Ego' method, while we use both static point correspondences and dynamic point correspondences for the `ST' method. `DLoss' denotes the discriminative loss function.
As can be seen, the results demonstrate that each component of our method has significantly improved prediction results. On Semantic-KITTI, our method leads to a performance boost of an absolute value of $4.8\%$ in \text{mIoU}, compared to the baseline's \text{mIoU} of $8.6\%$.

\begin{figure}
    \centering
    \subfloat{\begin{tikzpicture}  
    \node (image) {\includegraphics[width=0.22\textwidth]{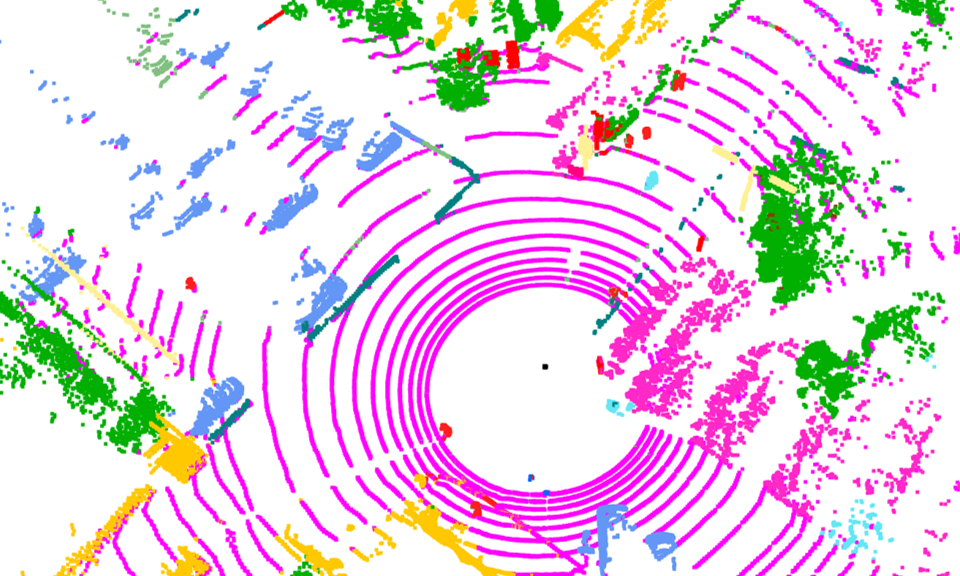}};
    \node [above = 0.0cm and 0.0cm of image] {Ground Truth};
    \node [left = 0.0cm and 0.0cm of image] {\rotatebox{90}{SemanticPOSS}};
    \end{tikzpicture}}
    \subfloat{\begin{tikzpicture} 
    \node (image) {\includegraphics[width=0.22\textwidth]{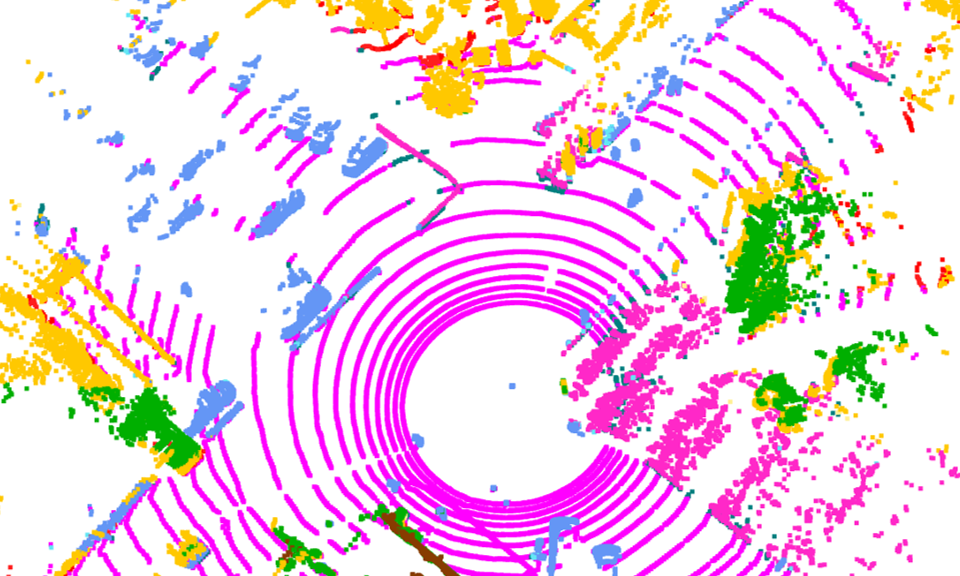}};
    \node [above = 0.0cm and 0.0cm of image] {Prediction};
    \end{tikzpicture}}
    \vfill
    \subfloat{\begin{tikzpicture} 
    \node (image) {\includegraphics[width=0.22\textwidth]{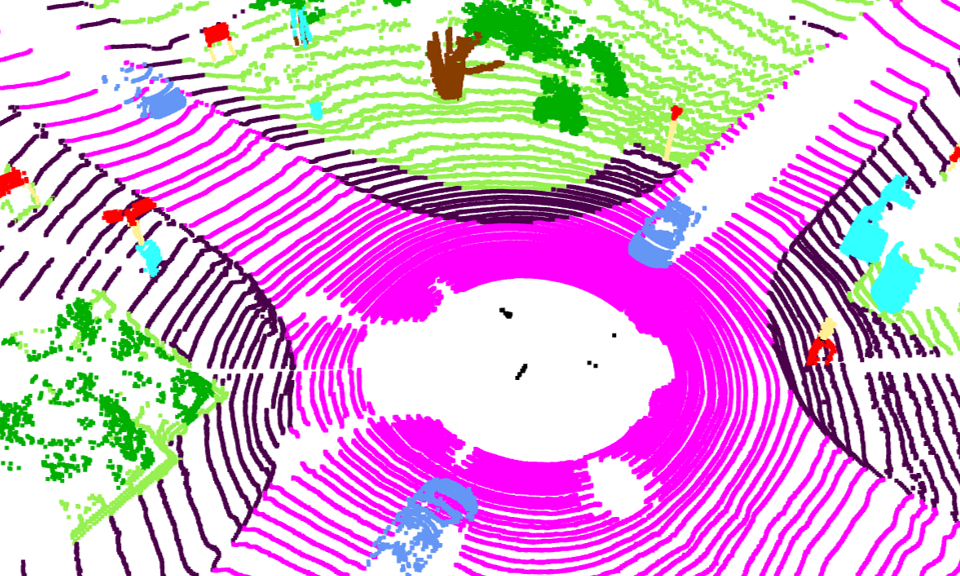}};
    \node [left = 0.0cm and 0.0cm of image] {\rotatebox{90}{Semantic-KITTI}};
    \end{tikzpicture}}
    \subfloat{\begin{tikzpicture}  
    \node (image) {\includegraphics[width=0.22\textwidth]{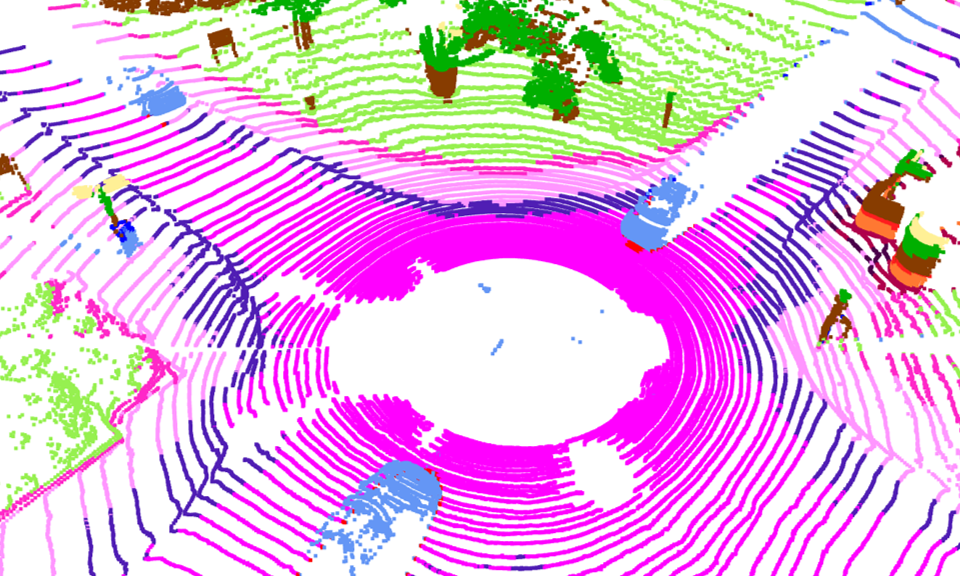}};
    \end{tikzpicture}}
    \vfill
    \subfloat{\begin{tikzpicture}  
    \node (image) {\includegraphics[width=0.22\textwidth]{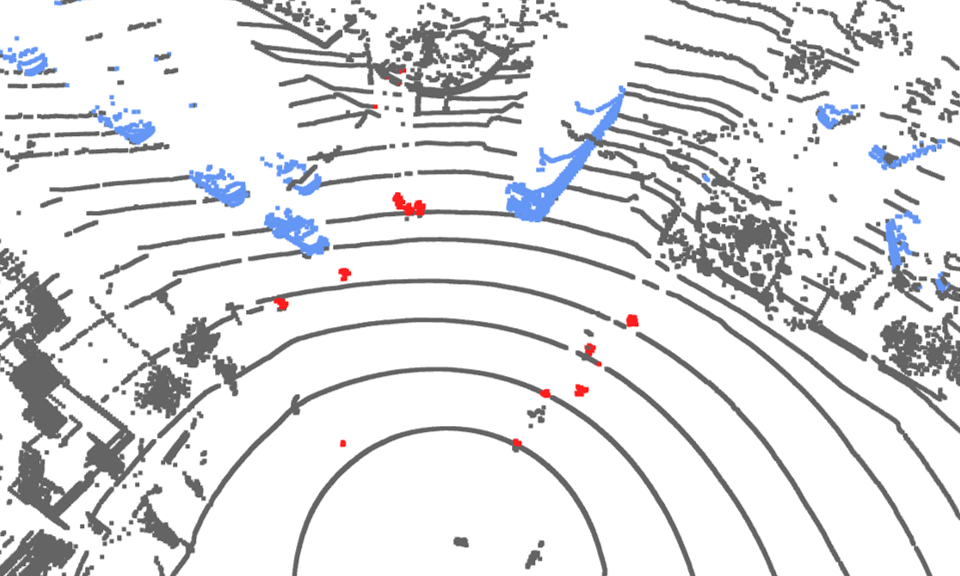}};
    \node [left = 0.0cm and 0.0cm of image] {\rotatebox{90}{FLORIDA}};
    \end{tikzpicture}}
    \subfloat{\begin{tikzpicture}  
    \node (image) {\includegraphics[width=0.22\textwidth]{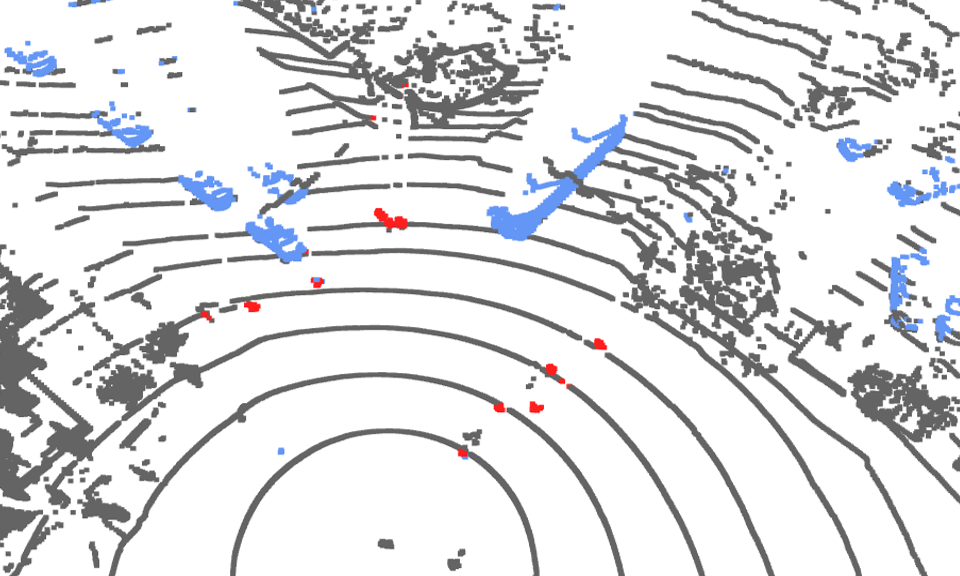}};
    \end{tikzpicture}}
    \caption{Ground truth and model prediction on SemanticPOSS dataset, Semantic-KITTI dataset, and FLORIDA dataset}
    \label{fig:main_result}
    
\end{figure}
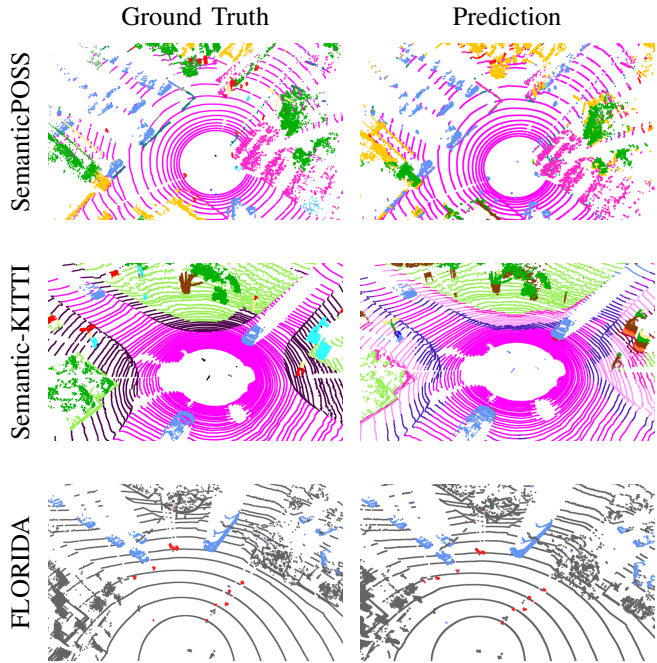

We evaluated methods on the SemanticPOSS dataset in Table~\ref{tab:semposs} and demonstrated that our method significantly enhances the \text{mIoU}. Our method has an absolute $9.1\%$ improvement when compared with the baseline's \text{mIoU} of $10.9\%$. Figure~\ref{fig:main_result}  provides qualitative results---our developed models efficiently segment roads, vegetation, cars, and other objects.


\subsection{Ablation Study}

\subsubsection*{Impact of Different Data Augmentations Approaches}
Data augmentation plays a crucial role in model learning: stronger augmentations could  further improve the model performance. Therefore,  we explored several effective point cloud augmentations including random translation, point flip along x-axis, random point down-sampling, and random rotation along the z-axis. For random rotation, we only rotated the point cloud by 180 degrees. The reason is that we found a 180-degree rotation gives better evaluation performance than arbitrary rotations. As shown in Table~\ref{tab:augmentation}, the results confirm the effectiveness of each data augmentation strategy. We use all data augmentation approaches for our developed models.

\subsubsection*{Impact of Frame Interval for Spatiotemporal Contrastive Learning}

Our approach creates spatiotemporal samples by selecting two adjacent frames with an frame interval. We set the interval to no larger than 30 frames (equivalent to a 3-second interval based on the LiDAR frame rate) to ensure that the two frames share sufficiently overlapped areas.   To derive the best interval, we tested different interval values from the set $F$. As shown in Table~\ref{tab:frame_interval},  for the `random' interval, we randomly selected a interval from $T$ for each pair. The interval $10$ and 
 the random interval perform better compared to other intervals. Our study suggests that the frame interval is an important hyperparameter and that a random interval may be a better strategy.




\begin{table}[t]
\centering
\caption{Ablation study of point cloud augmentations.}
\begin{tabular}{ c c c c c } 
\toprule
translation& 
point flip &
rotation& 
downsampling &
\text{mIoU}
\\
\midrule
\checkmark & - & - & - & 6.83\\
\checkmark & \checkmark & - & - & 8.58 \\
\checkmark & \checkmark & \checkmark & - & 9.82\\
\checkmark & \checkmark & \checkmark & \checkmark & 10.50 \\
\bottomrule
\end{tabular}

\label{tab:augmentation}
\end{table}
\begin{table}
\centering
\caption{Ablation study of frame interval for creating contrastive samples.}

\resizebox{0.48\textwidth}{!}{\begin{tabular}{ c   c c c c c c c c} 
\toprule
Frame Interval & 5 & 10 & 15 & 20 & 25 & 30 & Random 
\\
\midrule
\text{mIoU} & 13.8 & 15.2 & 14.3 & 14.0 & 13.8 & 10.2 & 15.1\\
\bottomrule

\end{tabular}}

\label{tab:frame_interval}
\end{table}
\begin{figure*}
    \centering
    \subfloat{\includegraphics[width=0.3\textwidth]{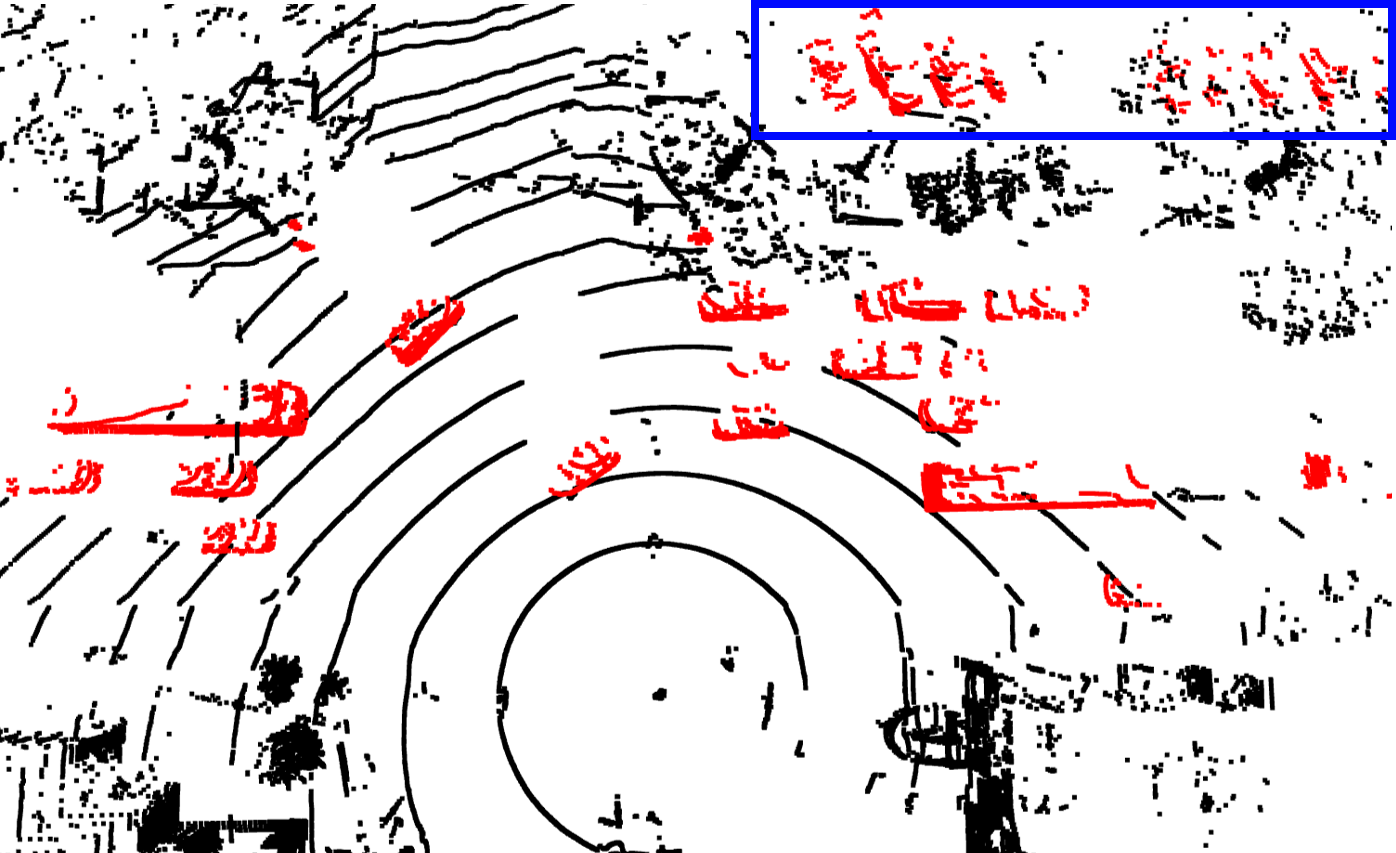}}
    \hfill
    \subfloat{\includegraphics[width=0.3\textwidth]{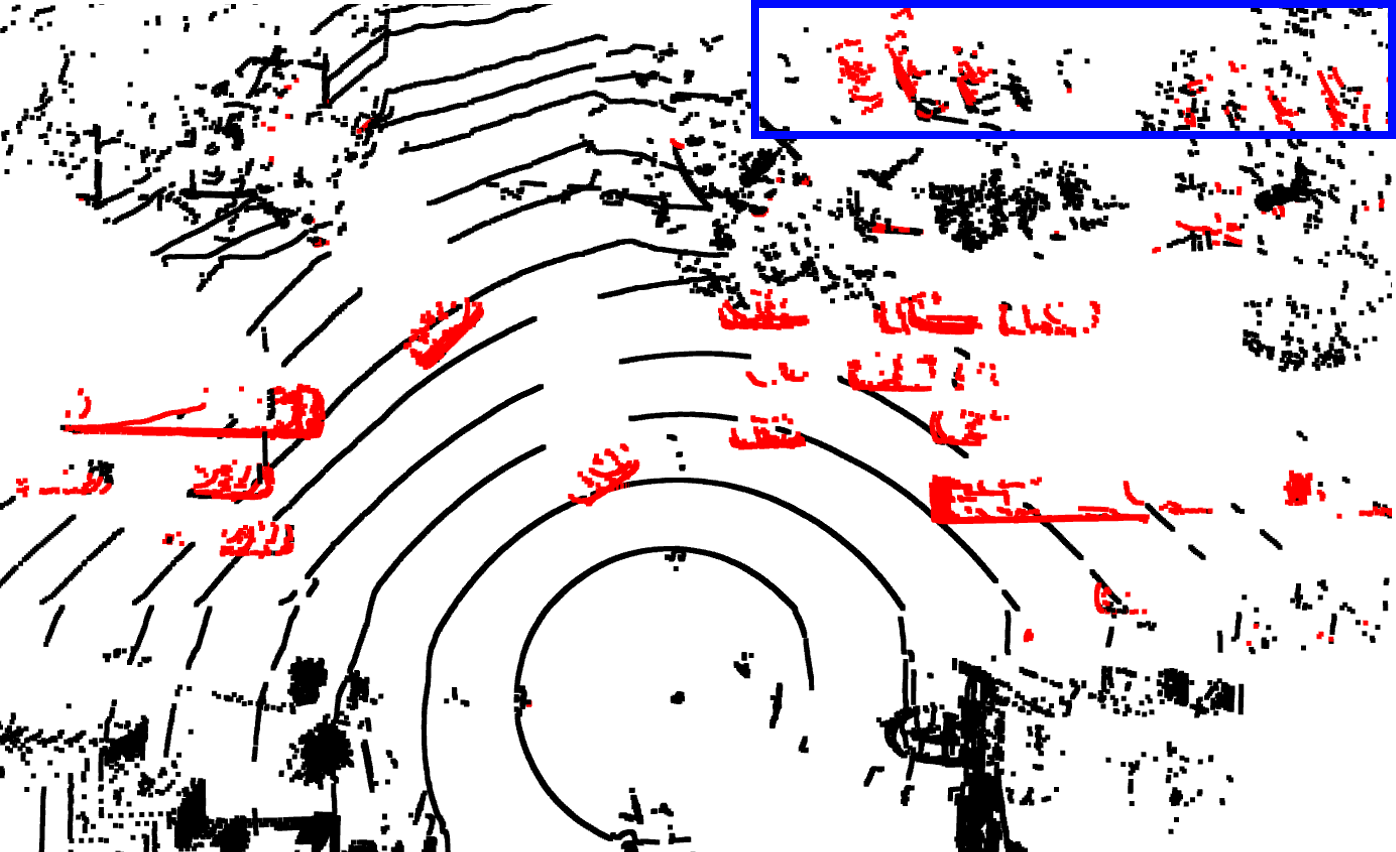}} 
    \hfill
    \subfloat{\includegraphics[width=0.3\textwidth]{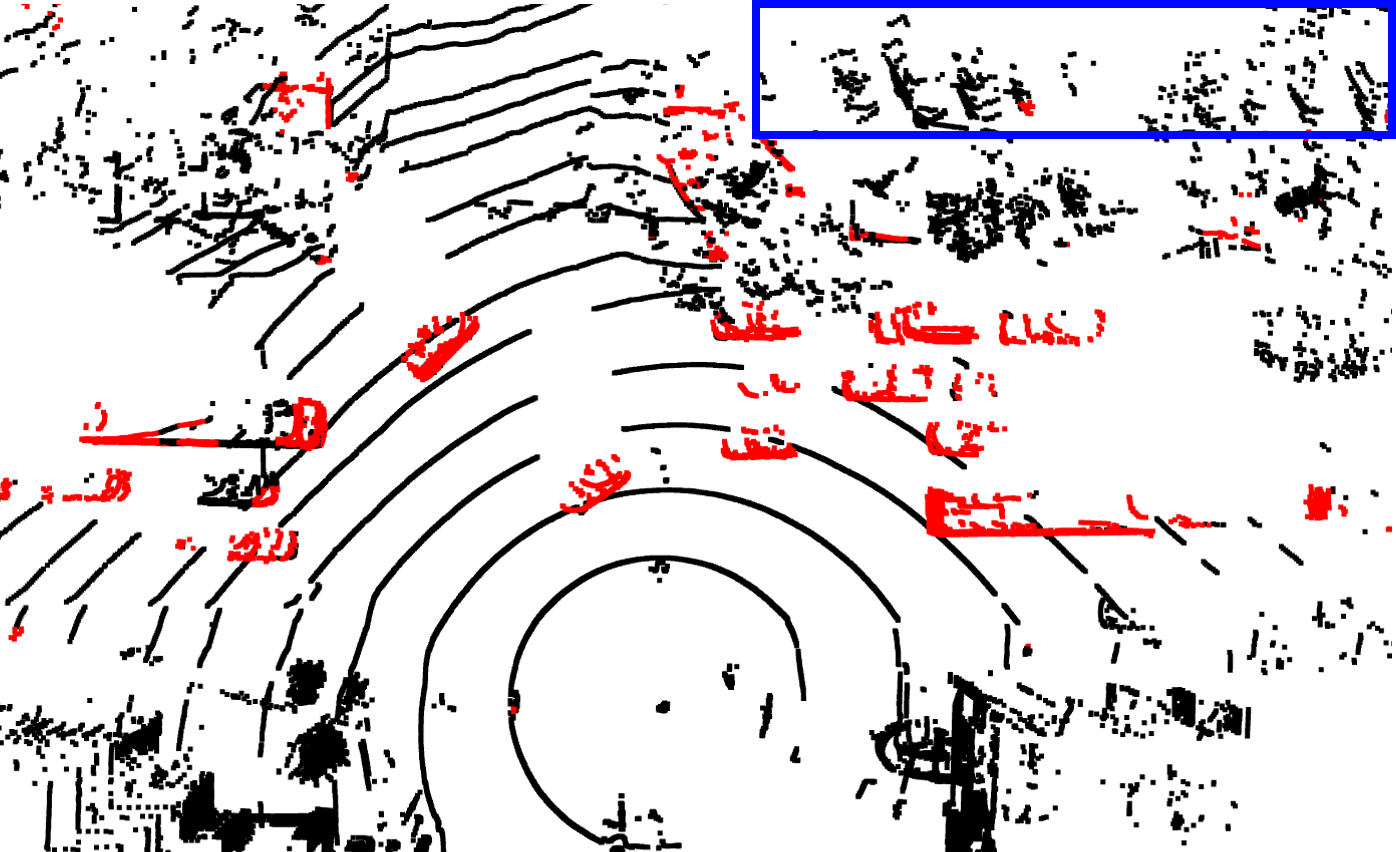}}
    \caption{Foreground-background segmentation: ground truth (left), heuristic method (middle), and simple threshold (right)}
    \label{fig:backgnd}
\end{figure*}

\subsubsection*{Generalization Capabilities}
We present the generalization capabilities on various LiDAR datasets collected from different environments. We tested our method on the  FLORIDA dataset collected from a traffic intersection \cite{wu2023florida}. Unlike the Semantic-KITTI dataset, collected from an onboard LiDAR mounted on a moving vehicle, the stationary LiDAR of the FLORIDA dataset provides an opportunity to analyze a complex, crowded, and safety-critical intersection containing a large number of pedestrians.  As most of the dataset's background is static, we only considered the segmentation performance of foreground objects, including static and dynamic objects. To support the training and evaluation, we bootstrapped the 3D bounding box-based tracking annotations of the dataset into point-wise semantic labels. We group all objects into three classes: Vehicles (consisting of cars, buses, and trucks), Pedestrians or People Class (consisting of pedestrians  and cyclists) and Background Class (consisting of non-movable objects like roads and buildings).

We address this segmentation task by decomposing it into two sub-problems: segmenting foreground and background points and classifying foreground points into semantic classes. The key motivation is to leverage the prior information---most background regions remain the same.  We seek to separate foreground and background points without ground truth labels by generating pseudo-labels for a binary segmentation model. We explored two method variants. The first one (`Simple Threshold') creates a pseudo-label using a simple dynamic score in Equation~\eqref{eq:dynamic2}: a point belongs to the foreground if it has a high dynamic score, otherwise it is in  the background. However, the major drawback is that such an approach cannot distinguish the points of parked cars as they are static but belong to the foreground. Therefore, we proposed the second method (`Heuristic') of classifying points into foreground, background, and uncertain groups. 

\begin{table}[t!]
\centering
\caption{Foreground-background segmentation on FLORIDA dataset.}
\begin{tabular}{ c  c c c c c c c c} 
\toprule
  & & Background & Foreground & mIoU 
\\
\midrule
\multirow{2}{5em}{Training Set}
&Simple Threshold & 95.6 & 54.4 & 75.0\\
&\text{Heuristic} & 97.0 & 68.7 & 82.8\\
\bottomrule

\multirow{2}{5em}{Test Set} 
&Simple Threshold& 96.5 & 44.3 & 70.4\\
&\text{Heuristic} & 97.9 & 63.7 & 80.8 \\
\bottomrule

\end{tabular}

\label{tab:florida_bng}
\end{table}

\begin{table}[t!]
\centering
\caption{Unsupervised semantic segmentation on Florida testing set using a single shot method and cascaded methods}
\begin{tabular}{ c   c c c c c c c c} 
\toprule
  & Vehicle & People & Background & mIOU  \\
\midrule 
\text{Single Shot} & 60.5 & 7.4 & 96.3 & 54.7\\
Dynamic-based Cascade & 47.0 &   24.3  &   96.4& 55.9 \\
\text{Heuristic}-based Cascade  &  69.1 &   38.0 &    98.1 &68.4\\

\bottomrule

\end{tabular}

\label{tab:florida_uss}
\end{table}We adopted the same approach as in Section~\ref{dynamicSeg} to identify dynamic and static points. The main modification is for the static points: we try to identify those points belonging to parked cars and classify them into the uncertain group. To do so, we run the DBSCAN algorithm on the static points and estimate the minimal bounding box of each clustering. For bounding boxes of a similar size to a car and close to the ground plane, we set their points as uncertain. 

We use the pseudo-label of foreground and background points for model training and ignore uncertain points. The backbone network used is SalsaNext with a minor modification to change the output size of the last layer to one. A sigmoid function is adopted to output $ \left ( 0,1 \right )$ values. The root mean square error (RMSE) loss is adopted as the objective function. We reported the FLORIDA results in Table~\ref{tab:florida_bng}. Compared to `Simple Threshold', i.e., requiring multiple frames to determine the dynamic score, `Heuristic' is advantageous because it only needs a single frame fed to a model to obtain a better foreground-background segmentation. `Heuristic' leads to a performance boost of an absolute $14.3\%$ and $19.4\%$ on  the FLORIDA training set and test set respectively. Figure~\ref{fig:backgnd} demonstrates the classification results. Cars within the blue rectangle are parked cars. `Simple Threshold' wrongly classifies all parked cars as background objects. In contrast, `Heuristic' based approach can effectively classify parked cars as foreground objects.




After foreground-background segmentation, we run the auto correspondence labeling algorithm proposed in Section~\ref{auto_correp} on foreground points, obtaining point-wise correspondences. We use a training strategy similar to training a semantic segmentation model for FLORIDA dataset as used for Semantic-KITTI dataset and SemanticPOSS dataset.  One difference is that we only feed foreground points to network and keep dynamic object correspondences.
We label this method `Heuristic-based Cascade'. For comparison, we also design two other experiments. For the first, we use the same training strategy as Semantic-KITTI and SemanticPOSS datasets, marked as `Single Shot'. The second one is similar to the `Heuristic-based Cascade' method except that the foreground-background segmentation results are generated by a simple threshold method, marked as `Dynamic-based Cascade'. The `Single Shot' method outperforms the `Dynamic-based Cascade' with regard to IoU of `Vehicle' but shows poor performance  on the `People' class, which indicates that a cascade approach is beneficial for tiny classes, i.e., people. The results (cf. Table~\ref{tab:florida_uss})  show that the `Heuristic-based Cascade' method outperforms the other two methods on all three categories, which demonstrates that a good foreground-background segmentation model can largely benefit the cascade approaches.







\section{CONCLUSIONS}\label{conclusion}
We have introduced a novel framework for addressing unsupervised segmentation in outdoor LIDAR sequences. A key contribution of our work is in learning the equivariance of point clouds by establishing spatiotemporal correspondences between point cloud pairs. To enhance the generalization capabilities of our method, we have further demonstrated a cascaded approach to tackle the unsupervised segmentation for traffic intersection scenarios. To the best of our knowledge, we are the first to perform unsupervised LiDAR semantic segmentation in both autonomous driving and infrastructure contexts. The experimental results demonstrate that our approach achieves competitive performance even when compared to some state-of-the-art supervised methods. 


\section*{Acknowledgments}
This work is supported by NSF CNS 1922782 and by the Florida Dept. of Transportation (FDOT), and FDOT District 5. The opinions, findings and conclusions expressed in this publication are those of the author(s) and not necessarily those of FDOT or the National Science Foundation.






\bibliographystyle{IEEEtran}
\bibliography{ref.bib}

\vfill

\end{document}